\documentclass[times, review, 10pt]{elsarticle}
\usepackage{amssymb}
\usepackage{setspace}
\usepackage{amsmath}
\usepackage{graphicx}
\usepackage{caption}
\usepackage{float}
\usepackage{booktabs}
\usepackage{multirow}
\usepackage{makecell}
\usepackage{bbding}
\usepackage[normalem]{ulem}
\useunder{\uline}{\ul}{}
\usepackage{adjustbox}
\usepackage[section]{placeins}
\usepackage[colorlinks=true, allcolors=blue]{hyperref}
\usepackage{lineno,hyperref}
\usepackage{threeparttable}
\usepackage{ulem}
\usepackage{color}
\journal{Nuclear Physics B}
\usepackage{cleveref}
\Crefname{Figure}{Fig}{Figures}
\begin{document}
\begin{frontmatter}
\title{IRDFusion: Iterative Relation-Map Difference guided Feature Fusion for Multispectral Object Detection}

\author[a]{Jifeng~Shen\corref{cor1}}
\cortext[cor1]{Corresponding author}
\ead{shenjifeng@ujs.edu.cn}

\author[a]{Haibo~Zhan}
\author[b]{Xin~Zuo}
\author[d]{Heng~Fan}
\author[d]{Xiaohui~Yuan}
\author[e]{Jun~Li}
\author[c]{Wankou~Yang}
\address[a]{School of Electrical and Information Engineering, Jiangsu University, Zhenjiang, 212013, China}
\address[b]{School of Computer Science and Engineering, Jiangsu University of Science and Technology, Zhenjiang, 212003, China}
\address[c]{School of Automation, Southeast University, Nanjing, 210096, China}
\address[d]{Department of Computer Science and Engineering, University of North Texas, Denton, TX 76207, USA}
\address[e]{School of Computing, Nanjing Normal University, Nanjing, 210046, Jiangsu}

\begin{abstract}
Current multispectral object detection methods often retain extraneous background or noise during feature fusion, limiting perceptual performance.
To address this, we propose an innovative feature fusion framework based on cross-modal feature contrastive and screening strategy, diverging from conventional approaches. 
The proposed method adaptively enhances salient structures by fusing object-aware complementary cross-modal features while suppressing shared background interference.
Our solution centers on two novel, specially designed modules: the Mutual Feature Refinement Module (MFRM) and the Differential Feature Feedback Module (DFFM). 
The MFRM enhances intra- and inter-modal feature representations by modeling their relationships, thereby improving cross-modal alignment and discriminative power.
Inspired by feedback differential amplifiers, the DFFM dynamically computes inter-modal differential features as guidance signals and feeds them back to the MFRM, enabling adaptive fusion of complementary information while suppressing common-mode noise across modalities.
To enable robust feature learning, the MFRM and DFFM are integrated into a unified framework, which is formally formulated as an Iterative Relation-Map Differential Guided Feature Fusion mechanism, termed IRDFusion. IRDFusion enables high-quality cross-modal fusion by progressively amplifying salient relational signals through iterative feedback, while suppressing feature noise, leading to significant performance gains.
In extensive experiments on FLIR, LLVIP and M$^3$FD datasets, IRDFusion achieves state-of-the-art performance and consistently outperforms existing methods across diverse challenging scenarios, demonstrating its robustness and effectiveness.
Code will be available at https://github.com/61s61min/IRDFusion.git.
\end{abstract}

\end{frontmatter}


\section{Introduction}
\label{sec1}
Multispectral object detection employs data from multiple spectral bands, such as visible and infrared light, for object recognition and localization. It is widely applied in autonomous driving and video surveillance tasks in poor weather conditions (e.g. darkness, fog, rain or snow). Compared to single-spectrum data, multispectral data can more comprehensively reflect the spectral characteristics of the object and its background, thereby significantly improving the robustness and accuracy of detection. It is worth noting that multispectral object detection is different from general multimodal object detection. Multispectral detection focuses on information captured from different spectral bands of the optical sensor system (e.g., RGB and infrared) \cite{liu2016multispectral, SHEN2024109913, shen2025multispectral}, where the modalities are physically correlated and often exhibit strong structural consistency. In contrast, multimodal detection usually involves heterogeneous sources such as images, texts, LiDAR point clouds, or audio \cite{hu2023fusionformer, li2022deepfusion}, where the modalities differ not only in physical characteristics but also in semantic representation, requiring more complex alignment and fusion strategies.

Although current multispectral object detection approaches have achieved significant progress by exploring cross-modal fusion strategies, several intrinsic limitations still remain. Modality-specific reconstruction methods (e.g., SCFR \cite{sun2024specificity}) try to preserve unique information, but they often overlook redundant background features that are simultaneously present in both modalities, thereby weakening the discriminability of fused representations. Transformer-based approaches (e.g., DAMSDet \cite{DAMSdet}, ICAFusion \cite{SHEN2024109913}) have attempted to capture global complementary information and address misalignment, but their heavy reliance on stacked attention blocks introduces high computational burden and excessive parameterization, which restricts scalability and real-time applicability. Alignment-driven strategies (e.g., CAGT \cite{yuan2024improving}) mitigate spatial misalignment at the region level, yet they are less effective in filtering modality-shared noise, leading to fused features that remain contaminated by background artifacts. 

\begin{figure}
    \centering
    \includegraphics[width=\linewidth]{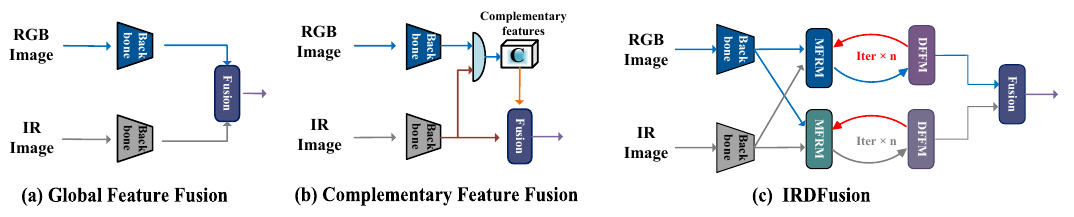}
    \caption{Comparison of the fusion structures between our method and existing methods}
    \label{comparision}
\end{figure}

These limitations collectively highlight two core challenges that remain insufficiently addressed: (1) how to effectively suppress modality-shared background interference while preserving salient complementary object features, and (2) how to achieve adaptive feature integration without incurring excessive complexity. To tackle the aforementioned challenges, we propose IRDFusion, a novel fusion framework inspired by the principles of differential operational amplifier circuits. As illustrated in \Cref{comparision}(c), IRDFusion introduces a iterative differential feedback mechanism that progressively guides feature fusion between RGB and thermal modalities. Specifically, the framework integrates two complementary modules: the Mutual Feature Refinement Module (MFRM) and the Differential Feature Feedback Module (DFFM).

The MFRM leverages attention to dynamically enhance semantics (e.g., object contours) while effectively suppressing background noise, thereby providing a stable and well-aligned semantic foundation for fusion. Building on this, the DFFM, inspired by differential amplifier circuits, computes inter-modal differences (e.g., color versus thermal radiation) and reinjects them into the fusion pipeline through iterative feedback loops. This mechanism preserves discriminative complementary cues and simultaneously filters redundant information.

The two modules operate in a closed-loop interaction: MFRM first consolidates modal features for reliable object localization, while DFFM refines modality-specific differences to enrich discriminative details. In turn, the refined differential cues enhance the alignment of modal features, creating a synergistic cycle that progressively amplifies salient signals and suppresses irrelevant background. As a result, IRDFusion produces fused representations with sharper contours and stronger discriminability. Besides, IRDFusion effectively mitigates the issues of weak feature alignment and loss of differential information observed in conventional fusion methods, thereby achieving superior detection accuracy in challenging scenarios such as occlusion and low-light conditions.

In summary, our main contributions are as follows:
\begin{itemize}
\item A Mutual Feature Refinement Module (MFRM) is proposed to enhance modal-specific features of object candidates between two modalities, ensuring robust feature alignment.

\item Inspired by the feedback differential amplifier circuits, a Differential Feature Feedback Module (DFFM) is proposed to calculate complementary discriminative features between the two modalities and simultaneously filters redundant information.
 
\item The MFRM and DFFM are jointly optimized to effectively integrate discriminative complementary information from different modalities through a dynamic differential relationship map feedback mechanism, which provides a new strategy for progressive multispectral feature fusion. 

\item The proposed method IRDFusion, building on MFRM and DFFM, achieves state-of-the-art performance on the FLIR, LLVIP and M$^3$FD datasets.
\end{itemize}

The rest of this paper is organized as follows: \Cref{sec2} reviews related work on multispectral object detection, summarizing existing methods and their advantages and limitations; \Cref{sec3} describes the details of our proposed method, including the model architecture and key techniques; \Cref{sec4} presents experimental results, comparing the performance of our method with existing approaches; \Cref{sec5} concludes the paper and discusses future research directions.

\section{Related Work}
\label{sec2}
\subsection{Object Detection}
\label{2.1}
Object detection is a fundamental task in the field of computer vision, primarily categorized into one-stage and two-stage detectors. One-stage detectors, such as YOLO \cite{7780460}, SSD \cite{liu2016ssd}, and RetinaNet \cite{8237586}, perform direct regression on feature maps, achieving high detection speeds. Methods like CenterNet \cite{zhou2019objects} and DETR \cite{DETR2020} further simplify the detection pipeline by directly regressing object center points or employing Transformers for end-to-end detection. In contrast, two-stage detectors, such as R-CNN \cite{girshick2014rich}, SPPNet \cite{he2015spatial}, and FPN \cite{lin2017feature}, first generate candidate regions and then perform refined classification and bounding box regression, typically achieving higher accuracy. Moreover, detection methods can be divided into anchor-based and anchor-free approaches. Anchor-based methods, such as YOLO and RetinaNet, rely on predefined anchor boxes for object prediction, while anchor-free methods, such as CenterNet and FCOS \cite{tian2020fcos}, locate object center points or boundary points directly, reducing reliance on anchor box design and lowering computational complexity. Recent improvements on the DETR framework, such as DINO \cite{zhang2022dino}, further enhance performance and training efficiency through contrastive denoising training and improved query selection. Similarly, MADet \cite{xie2023mutual} introduces the concept of mutual-assistance learning by reintegrating decoupled classification and regression features and jointly leveraging anchor-based and anchor-free regression.
In our research, we select the DETR framework due to its end-to-end training capability, simplified detection pipeline, and effective global context modeling, which enhances detection performance, especially in complex scenes.
\subsection{Multispectral Feature Fusion for Detection}
\label{2.2}
Multispectral Object Detection combines RGB and thermal modalities to improve detection performance in complex scenarios. Early studies such as ConvNet \cite{liu2016multispectral} introduce a multispectral pedestrian dataset and significantly reduce detection errors through an ACF-based extension method. IAF R-CNN \cite{li2019illumination} incorporates an illumination-aware mechanism and a multitask learning framework, enhancing robustness under varying lighting conditions.

In terms of feature alignment and modality fusion, AR-CNN \cite{zhang2019weakly} proposes a region feature alignment module and features reweighting method to address weak alignment, improving multimodal fusion. Similarly, MCHE-CF \cite{li2023multiscale} introduces multiscale homogeneity enhancement and confidence fusion, improving modality complementarity through a channel attention mechanism. LG-FAPF \cite{cao2022locality} improves discrimination and fusion by aggregating features with locality guidance and pixel-level fusion. MMA \cite{shen2022mask} proposes an explicit features modulation method guided by masks, significantly improving the performance of multi-task learning in object detection and box-level segmentation, especially under scale variations and occlusion.  SPA and AFA \cite{zuo2023improving} improve the quality of feature fusion in multispectral pedestrian detection through the scale-aware permutated attention mechanism with adjacent-branch feature aggregation module, effectively reducing the miss rate for small pedestrians.

Despite these advances, many fusion methods still face two core challenges, echoing insights from oriented object detection: effectively suppressing modality-shared background while preserving salient complementary object features, and achieving adaptive feature integration without excessive complexity. Analogous to how DFDet \cite{xie2024oriented} leverages contextual priors and SFRNet \cite{cheng2023sfrnet} refines discriminative features for oriented detection, multispectral fusion can benefit from task-specific module designs that explicitly model both commonality and complementarity between modalities.

In the context of Transformer and attention-based fusion methods, CFT \cite{qingyun2021cross} utilizes self-attention mechanisms for inter-modality interaction, while ICAFusion \cite{SHEN2024109913} employs iterative cross-attention to reduce model complexity and improve performance. AMSF \cite{bao2022attention} and TFDet \cite{zhang2024tfdet} introduce object awareness and attention mechanisms, significantly improving detection accuracy and background suppression. For redundant information suppression, RISNet \cite{Improving2020} minimizes redundancy between RGB and IR images, optimizing complementary fusion and improving detection performance. LGADet~\cite{zuo2023lgadet} improves inference speed by utilizing a lightweight backbone and an anchor-free detection framework and employs a hybrid attention mechanism to enhance accuracy.

Finally, PIAFusion \cite{tang2022piafusion} proposes a progressive image fusion network based on illumination awareness, which adaptively optimizes infrared and visible light image fusion under varying lighting conditions while preserving object and texture details. GAFF \cite{zhang2021guided} employs attention mechanisms to dynamically weigh and fuse multispectral features, significantly improving detection accuracy with low computational cost.

In contrast, our proposed IRDFusion model introduces a novel relational differential feedback mechanism for feature fusion. Specifically, IRDFusion first strengthens semantic information across modalities, while simultaneously emphasizing discriminative differential cues. It then extracts and feeds back inter-modal differences as guidance signals, thereby amplifying complementary object features and suppressing redundant background information. Through this iterative feedback process, IRDFusion progressively refines cross-modal alignment, leading to enhanced accuracy and robustness compared to existing fusion approaches.
\section{The Proposed Method}
\label{sec3}
\subsection{Architecture}
\label{3.1}
\begin{figure}[!t]
    \centering
    \includegraphics[width=1.0\linewidth]{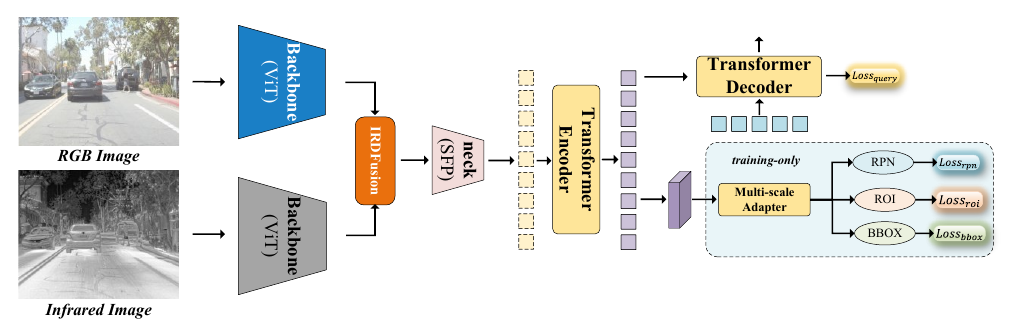}
    \caption{The architecture of the model.}
    \label{architecture}
\end{figure}
The model first employs a dual-branch backbone to extract features from both RGB and thermal modalities, while the proposed IRDFusion module is utilized to progressively fuse cross-modal features. IRDFusion enhances feature representations by amplifying inter-modal differences and leveraging them as guidance signals to steer the fusion process step by step. The fused representations are subsequently processed by a Simple Feature Pyramid (SFP) neck \cite{zhou2024optimizing}, followed by a Transformer Encoder, and finally fed into the multiple parallel detection heads of Co-DETR. The detection head design remains consistent with Double-Co-DETR \cite{zhou2024optimizing}. This architecture effectively integrates complementary cross-modal cues, leading to substantially improved detection performance under challenging conditions.
\begin{figure}[!t]
    \centering
    \includegraphics[width=1.0\linewidth]{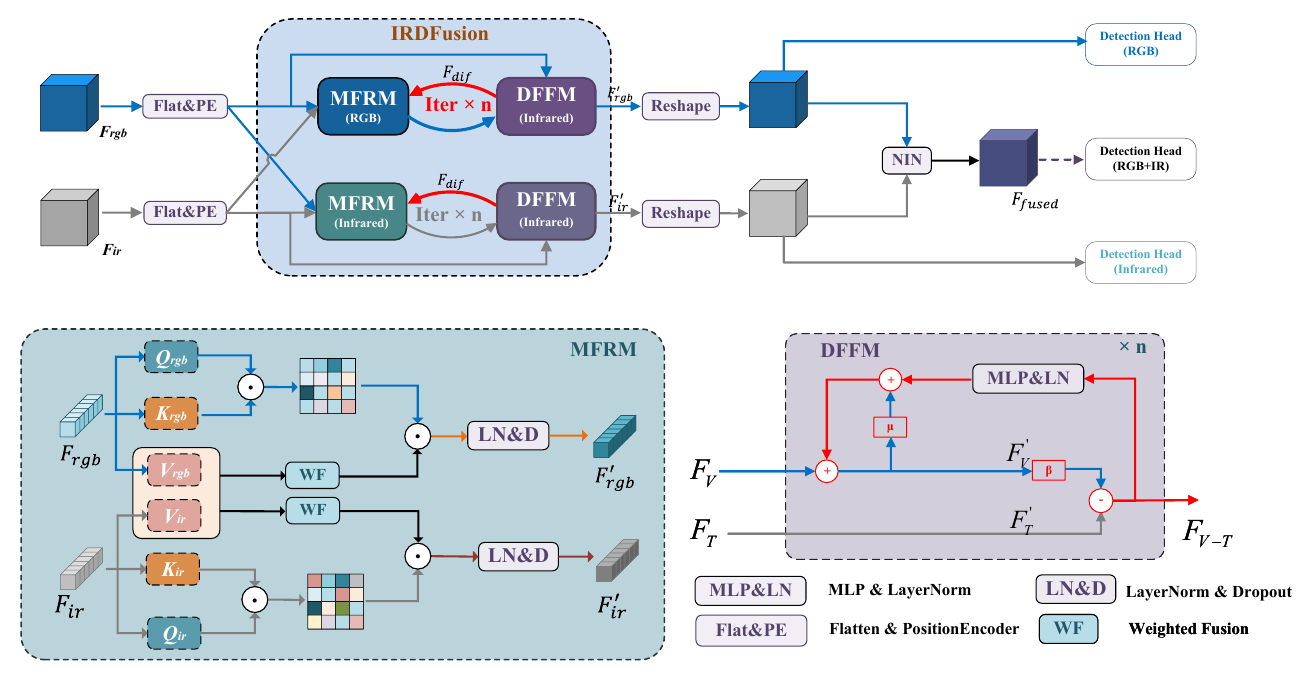}
    \caption{Illustration of the proposed IRDFusion module. The red data stream is DFFM. In this context, $LN\&D$ refers to Layer Normalization (LN) and Dropout, $MLP\&LN$ refers to the MLP layer followed by Layer Normalization, $ \lambda, \beta $ and $ \mu $ are learnable parameters. $Flat\&PE$ refers to organizing feature maps into sequences and adding positional encoding information and $Reshape$ refers to converting the sequence back into a feature map.}
    \label{IRD}
\end{figure}
\subsection{Mutual Feature Refinement Module (MFRM)}
\label{3.2}
The Mutual Feature Refinement Module (MFRM) is designed to enhance the feature representations between two modalities, thereby improving cross-modal consistency and discriminative capability. Its core idea is to leverage the self-attention matrix of a single modality to interact with the weighted Value features of both modalities within a Transformer structure. In this way, MFRM amplifies cross-modal representations and produces more informative fused features. Specifically, as illustrated in \Cref{IRD}, the features of the two modalities are first projected through distinct weight matrices $W$ to generate Query, Key, and Value matrix. These vectors are then processed through self-attention, as described in Eq.~\eqref{attenqkv}, to obtain the corresponding attention matrices $A_i$, $i \in \{v, t\}$ for each modality.
\begin{eqnarray}
	\centering
	\begin{array}{l}
		\left\lbrack {Q_{i},K_{i},V_{i}}\right\rbrack = F_{i} \cdot \left\lbrack {W^q_{i}, W^k_{i}, W^v_{i}}\right\rbrack \\[2mm]
        A_{i} = Softmax  \frac{\left( ~Q_{i} \cdot K_{i}~\right)} {\sqrt{d}}
	\end{array}
	\label{attenqkv}
\end{eqnarray}
where $F_i$ represents the input features of the RGB or IR modalities, respectively. $Q_i, K_i, V_i$ denote the query, key, and value matrices, $\cdot$ represents matrix multiplication, while $ W^q_i, W^k_i, W^v_i$ are the weight matrices for linear transformation and $A_{i}$ represents the attention matrix, and $d$ represents the feature dimension.\par
Secondly, taking the RGB branch as an example, we integrate information from the infrared (IR) modality by incorporating the Value vector $V_{t}$ from the IR branch. Specifically, the attention matrix of the RGB branch is applied to the IR value vector, enabling the model to emphasize infrared cues within the spatial regions that the RGB branch attends to. In this way, the information between the two modalities is effectively reinforced, thereby enhancing their interaction and fusion. This process is formally expressed in Eq.~\eqref{fuse_v}:
\begin{eqnarray}
	\centering
	\begin{array}{l}
        V_{f\_v} = V_{v} + \lambda_{v} \cdot V_{t} \\[2mm]
        V_{f\_t} = V_{t} + \lambda_{t} \cdot V_{v}
	\end{array}
	\label{fuse_v}
\end{eqnarray}
where \( V_{f\_v} \) and \( V_{f\_t} \) are the fused Value features of the RGB and IR modalities.\par
During the fusion of the Value vectors, we introduce a learnable parameter in Eq.~\eqref{lambda}, drawing inspirations from \cite{ye2024differential}. This parameter allows the model to adaptively adjust the fusion process, improving robustness by enabling the model to scale the feature fusion based on the characteristics of the input data. This adaptive mechanism contributes to improved performance and greater flexibility in feature alignment.
\begin{eqnarray}
	\centering
	\begin{array}{l}
        \lambda_i = \exp \left( \lambda_{q1} \cdot \lambda_{k1} \right) - \exp \left( \lambda_{q2} \cdot \lambda_{k2} \right) + \lambda_{init}, i \in \{v, t \}
	\end{array}
	\label{lambda}
\end{eqnarray}
where \( \lambda_{v} \) and \( \lambda_{t} \) are the fusion weights for the modalities, controlled by learnable vectors \( \lambda_{q1}, \lambda_{q2}, \lambda_{k1}, \lambda_{k2} \) and the initial weight \( \lambda_{init} \).

Finally, we obtain the cross-modal amplified features based on Eq.~\eqref{MFRM}.
\begin{eqnarray}
    \centering
    \begin{array}{l}
        F_{i}^{'} =  A_{i} \cdot V_{f\_i}, i \in \{v, t \}
    \end{array}
    \label{MFRM}
\end{eqnarray}
where \( F'_{i} \) is the final fused features, and $V_{f\_i} $ means fused Value.
\subsection{Differential Feature Feedback Module (DFFM)}
\label{3.3}
The Differential Feature Feedback Module (DFFM) is inspired by differential feedback amplifier circuits, and is designed to leverage inter-modal differential features as guidance signals for dynamic cross-modal fusion. Specifically, these differential features capture the non-overlapping information between RGB and IR modalities, thereby highlighting their complementary characteristics while suppressing shared background noise. As illustrated in the lower part of \Cref{IRD}, taking the RGB branch as an example, the differential features between the RGB and IR modalities are first computed, with a learnable parameter $\beta$ introduced to adaptively control their contribution. The resulting differential features are then weighted and fed back into the RGB features, amplifying inter-modal difference signals and guiding the MFRM in extracting discriminative cues from the other modality. Through iterative feedback, the DFFM progressively enhances complementary information while filtering redundant noise, leading to more robust and adaptive cross-modal representations. The process, taking the RGB branch as an example, is formulated in Eq.~\eqref{DFFM}:
\begin{eqnarray}
	\centering
	\begin{array}{l}
		F^{(k)}_{dif\_v} = F^{'(k)}_{t} - \beta \cdot F^{'(k)}_{v} \\[2mm]
        F^{(k+1)}_{v} = \mu \cdot F^{(k)}_{v} + \alpha \cdot MLP\left(LN\left(F^{(k)}_{dif\_v}\right) \right) \\[2mm]
        F^{(k+1)}_{t} = F_{t} 
	\end{array}
	\label{DFFM}
\end{eqnarray}
where \(\alpha\) , \(\beta\) and \(\mu\) are learnable parameters. $MLP$ and $LN$ denote MLP layer and Layer Normalization, respectively. $F^{(k)}_{dif\_v}$ represents the difference features of the IR modality relative to the RGB modality. $F^{'(k)}_{i}$ refers to the output features of the \(k\)-th iteration of MFRM layer , while $F^{(k+1)}_{i}$ denotes the input features for the \((k+1)\)-th iteration of MFRM layer.

\subsection{Refactor into Iterative Relation Map Difference guided framework}
To enable robust feature learning, the MFRM and DFFM are integrated into a unified framework, which is formally formulated as an Iterative Relation-Map Differential Guided Feature Fusion framework, named IRDFusion. 
After feature refinement of both $F_{v}^{'}$ and $F_{t}^{'}$ from MFRM, the goal of $F_{v-t}$ is to obtain object-aware complementary feature across modality and eliminate common-mode background and noise information.
Following the Eq.~\eqref{attenqkv}-~\eqref{DFFM}, the cross-modal differential feature $F_{v-t}$ can be reformulated in Eq.~\eqref{f_vt}. 
Through some formula substitutions, the reconstructed feature with attention map $C_{(v-t)2v}$ represents the relation map difference between RGB and IR attention maps, and $F_{v-t}$ can also be considered as a difference of reconstructed feature from Value feature between visible and thermal branches. Taking visible modality as an example, the cross-modal differential feature $F_{v-t}$ is employed to feedback in a progressive way to refine features from $F_{v}^{(k+1)}$, where $k$ denotes the index of iteration, as shown in Eq.~\eqref{f_v_update}.
\begin{align}
F_{v}^{'} &= \text{softmax}\!\left(\frac{Q_{v}K_{v}^\top}{\sqrt{d}}\right)(V_{v}+\lambda_{v}V_{t}) 
           = A_{v2v}(V_{v}+\lambda_{v}V_{t}) \\[2mm]
F_{t}^{'} &= \text{softmax}\!\left(\frac{Q_{t}K_{t}^\top}{\sqrt{d}}\right)(V_{t}+\lambda_{t}V_{v}) 
           = A_{t2t}(V_{t}+\lambda_{t}V_{v}) \\[2mm]
F_{v-t} &= F_{v}^{'} - \beta F_{t}^{'} \notag \\
        &= A_{v2v}(V_{v}+\lambda_{v}V_{t}) - \beta A_{t2t}(V_{t}+\lambda_{t}V_{v}) \notag \\
        &= (A_{v2v} - \beta \lambda_{t}A_{t2t})V_{v} - (\beta A_{t2t}- \lambda_{v} A_{v2v})V_{t}\notag \\
        &= C_{(v-t)2v}V_{v} - C_{(t-v)2t}V_{t} \label{f_vt} \\[2mm]
F_{v}^{(k+1)} &= F_{v-t}^{(k)} + \beta F_{t}^{'} \label{f_v_update}
\end{align}
\noindent where $A_{v2v}$ and $A_{t2t}$ denote the relationship of intra-modal features for visible and thermal branches respectively. It is worth mentioning that the thermal image feature $F_{t}$ is fixed during the feature refinement for the RGB image branch. The refinement of thermal image branch $F_{t}$ is similar to Eq.~\eqref{f_v_update}, which is omitted here for clarity.

\subsection{Loss function}
\label{3.4}
In this work, we adopt the CoDetr loss, as in \cite{zhou2024optimizing}, for training. The CoDetr loss function integrates multiple components to optimize classification and localization performance. The main detection head (CoDINOHead \cite{zhang2022dino}) uses Quality Focal Loss for classification, effectively addressing class imbalance issues, and uses L1 Loss and GIoU Loss for bounding box regression and localization accuracy, respectively.

In addition to the main detection head, CoDetr also includes three auxiliary detection heads. The RPN Head \cite{girshick2015fast} applies Cross Entropy Loss for object-background classification and utilizes L1 Loss to refine bounding box proposals. The ROI Head \cite{girshick2015fast} adopts Cross Entropy Loss for category prediction and employs GIoU Loss to improve the precision of bounding box regression. The Bbox Head \cite{girshick2015fast} utilizes Focal Loss for classification, GIoU Loss for regression, and Cross-Entropy Loss for centerness prediction, contributing to improved detection accuracy.

This comprehensive loss design strikes a balance between robust classification and precise localization. The auxiliary detection heads complement the main detection head, further improving overall detection performance.
\section{Experiments}
\label{sec4}

\subsection{Implementation details}
\label{subsec4.2}
We use the double-co-detr framework from \cite{zhou2024optimizing} for our experiments. All experiments are carried out using PyTorch on a system equipped with an Intel i7-9700 CPU, 64 GB of RAM, and a Nvidia RTX 3090 GPU (24 GB of memory). The image input size is set to $640 \times 640$, and the data augmentation is followed by the v1 version from \cite{zhou2024optimizing}, with all other settings remaining consistent with those in the original paper. In the final experiments, the FLIR and LLVIP datasets are trained for 12 epochs, while the M$^3$FD dataset is trained for 36 epochs. In the ablation studies, NiNfusion is used as the baseline, and all experiments are trained for 12 epochs with a batch size of 1 with the same settings. Our code and model will be released for reproducing our results.

\subsection{Dataset and evaluation metric}
\textbf{FLIR} \cite{FLIR}: FLIR is a high-quality dataset for multispectral object detection, consisting of paired IR and RGB images. It is primarily used in autonomous driving and surveillance applications. FLIR includes object categories such as 'person', 'car', and 'bicycle'. With detailed annotations, it is particularly suitable for research on cross-modal fusion between infrared and visible light modalities.

\textbf{LLVIP} \cite{jia2021llvip}: LLVIP focuses on pedestrian detection in low-light conditions. LLVIP contains paired RGB and IR images and is specifically designed to address the challenges of multispectral pedestrian detection in low-light scenarios. It is widely used for research on multimodal data fusion methods in low-illumination environments.

\textbf{M$^3$FD} \cite{liu2022target}: It includes 4,200 infrared and visible light aligned image pairs collected under various environments such as different lighting, seasons, and weather scenarios. It encompasses six typical categories in automated driving and road surveillance. M$^3$FD is partitioned according to an 8:2 training/testing split as provided in \cite{Liang2023explicit}.\par
\textbf{Mean Average Precision (mAP)}:
The COCO mAP (mean average precision) evaluation metric is a standard for assessing object detection models. The mean AP is calculated across all classes and IoU thresholds (from 0.5 to 0.95, in increments of 0.05), offering a comprehensive evaluation of detection accuracy and localization precision. mAP50 measures AP at IoU = 0.5, focusing on sufficient overlap, while mAP75 evaluates AP at IoU = 0.75, requiring stricter localization.
\label{subsec4.1}

\subsection{State-of-the-art Comparison}
\label{subsec4.3}
\subsubsection{Comparison on the FLIR dataset}
\label{subsubsec4.3.1}
\begin{table}[!t]
\centering
\caption{Comparison on the FLIR dataset. The bold text represents the best result, and the underlined text represents the second best result. ‘-’ is used to indicate papers that have been published on arXiv but have not been officially published yet.}
\resizebox{0.65\columnwidth}{!}{%
\begin{tabular}{@{}rccccc@{}}
\toprule
Methods                            & Year & Source & mAP50         & mAP75         & mAP         \\ \midrule
GAFF \cite{zhang2021guided}        & 2021 & CVPR   & 72.9          & 32.9          & 36.6          \\
CFT \cite{qingyun2021cross}        & 2021 & -      & 78.7          & 35.5          & 40.2          \\
AMSF-net \cite{bao2022attention}   & 2022 & PRCV   & 78.9          & 35.5          & 40.9          \\
MFPT \cite{zhu2023multi}           & 2023 & TITS   & 80.0          & -             & -             \\
LRAF-Net \cite{fu2023lraf}         & 2023 & TNNLS  & 80.5          & -             & 42.8          \\
ICAFusion \cite{SHEN2024109913}    & 2024 & PR     & 79.2          & 36.9          & 41.4          \\
MMFN \cite{yang2024multidimensional}& 2024 & TCSVT  & 80.8          & -             & -          \\
RSDet \cite{zhao2024removal}       & 2024 & -      & 81.1          & -             & 41.4          \\
UniRGB-IR \cite{yuan2024unirgb}    & 2024 & -      & 81.4          & 40.2          & 44.1       \\
YOLOXCPCF \cite{hu2024rethinking}  & 2024 & TITS   & 82.1          & 41.2          & 44.6       \\
SCFR  \cite{sun2024specificity}    & 2024 & TITS   & 82.3          & 35.7          & -       \\
GM\_DETR \cite{xiao2024gm}         & 2024 & CVPR   & 83.9          & 42.6          & 45.8       \\
MMPedestron \cite{zhang2024pedestrian} & 2024 & ECCV      & 86.4          & -          & -      \\
DAMSDet \cite{DAMSdet}             & 2024 & ECCV   & {\ul 86.6}    & \textbf{48.1} & {\ul 49.3} \\ 
Fusion-Mamba \cite{dong2025fusion} & 2025 & TMM      & 84.9          & 45.9          & 47.0          \\\midrule
Baseline                           & 2025    & -      & 84.8          & 44.0          & 46.9          \\
Ours                               & 2025 & -      & \textbf{88.3} & {\ul 48.0}    & \textbf{50.7}    \\ \bottomrule
\end{tabular}%
\label{FLIR}
}
\end{table}

As shown in \Cref{FLIR}, our method achieves the best mAP50 while maintaining competitive mAP75, improving the baseline by 3.5\% and surpassing DAMSDet by 1.7\%, demonstrating its superior detection capability at lower IoU thresholds. For mAP75 and overall mAP, although slightly below the best-performing method, our approach still improves the baseline by 4.0\% and 3.8\%, respectively, indicating robust performance across stricter evaluation criteria. These gains stem from the core design of IRDFusion: the MFRM reinforces semantic features while suppressing redundant background, and the DFFM dynamically extracts and feeds back inter-modal differences to guide fusion. This iterative feedback progressively amplifies complementary object information and filters common-mode noise, enhancing both cross-modal alignment and discriminative power. 
\subsubsection{Comparison on the LLVIP dataset}
\label{subsubsec4.3.2}
\begin{table}[!t]
\centering
\caption{Comparison on the LLVIP dataset. The bold text represents the best result, and the underlined text represents the second best result.}
\resizebox{0.7\columnwidth}{!}{%
\begin{tabular}{@{}rccccc@{}}
\toprule
Methods                                 & Year & Source & mAP50         & mAP75        & mAP     \\ \midrule
CFT \cite{qingyun2021cross}             & 2021 & -      & 97.5          & 72.9          & 63.6          \\
CSAA \cite{cao2023multimodal}           & 2023 & CVPR   & 94.3          & 66.6          & 59.2          \\
LRAF-Net \cite{fu2023lraf}              & 2023 & TNNLS  & 97.9          & -             & 66.3          \\
UniRGB-IR \cite{yuan2024unirgb}         & 2024 & -      & 96.1          & 72.2          & 63.2       \\
YOLOXCPCF \cite{hu2024rethinking}       & 2024 & TITS   & 96.4          & 75.4          & 65.2       \\
AMSF-net \cite{bao2022attention}        & 2022 & PRCV   & 97.0          & 74.0          & 64.5          \\
MMFN \cite{yang2024multidimensional}    & 2024 & TCSVT  & 97.2          & -             & -          \\
GM\_DETR \cite{xiao2024gm}              & 2024 & CVPR   & 97.4          & {\ul 81.4}    & {\ul 70.2}       \\
SCFR  \cite{sun2024specificity}         & 2024 & TITS   & 97.5          & -          & -       \\
MS\_DETR \cite{xing2024ms}              & 2024 & TITS   & 97.9          & 76.3          & 66.1          \\
DAMSDet \cite{DAMSdet}                  & 2024 & ECCV   & 97.9          & 79.1          & 69.6     \\ 
ICAFusion \cite{SHEN2024109913}         & 2024 & PR     & \textbf{98.4} & 76.2           & 64.5 \\ 
Fusion-Mamba \cite{dong2025fusion}      & 2025 & TMM   & 97.0          & -             & 64.3           \\\midrule
Baseline                                & 2025    & -      & { \ul 98.0}   & 80.7          & 69.5          \\
Ours                                    & 2025 & -      & \textbf{98.4} & \textbf{83.1} & \textbf{70.9} \\ \bottomrule
\end{tabular}%
\label{LLVIP}
}
\end{table}
As shown in \Cref{LLVIP}, our method also achieves the best performance across all three metrics, with improvements of 0.4\%, 2.4\%, 1.4\% in mAP50, mAP75 and overall mAP compared to the baseline, and gains of 0.5\%, 4.0\%, 1.3\% over DAMSDet \cite{DAMSdet}. These improvements demonstrate the effectiveness of IRDFusion in pedestrian detection: by reinforcing semantic structures and leveraging differential cues, the framework enhances feature alignment and preserves discriminative information, particularly under challenging conditions such as occlusion or crowding, leading to more accurate detection.
\subsubsection{Comparison on the M\texorpdfstring{$^{3}$}{}FD dataset}
\label{subsubsec4.3.3}
\begin{table}[!t]
\centering
\caption{Comparison on the M$^3$FD dataset. The bold text represents the best result, and the underlined text represents the second best result.}
\resizebox{0.55\columnwidth}{!}{%
\begin{tabular}{@{}rcccc@{}}
\toprule
Methods                                 & Year & Source & mAP50         & mAP           \\ \midrule
TarDAL \cite{liu2022target}             & 2022 & CVPR   & 80.5          & 54.1          \\
CDDFusion \cite{zhao2023cddfuse}        & 2023 & CVPR   & 81.1          & 54.3          \\
IGNet \cite{li2023learning}             & 2023 & MM     & 81.5          & 54.5          \\
DAMSDet \cite{DAMSdet}                  & 2024 & ECCV   & 80.2          & 52.9          \\
MMFN \cite{yang2024multidimensional}    & 2024 & TCSVT  & 86.2          & -             \\
ICAFusion \cite{SHEN2024109913}         & 2024 & PR     & \textbf{90.8} & { \ul 60.9} \\
Fusion-Mamba \cite{dong2025fusion}      & 2025 & TMM   & {\ul88.0}     & \textbf{61.9} \\ \midrule
Baseline                                & 2025    & -      & 87.1          & 58.2    \\
Ours                                    & 2025 & -      & \textbf{90.8} & \textbf{61.9} \\ \bottomrule
\end{tabular}%
\label{M$^3$FD}
}
\end{table}
As presented in \Cref{M$^3$FD}, our method attains the highest performance across all metrics, matching the best model in overall mAP while achieving a 2.8\% improvement in mAP50. Compared to the baseline, both metrics increase by 3.7\%. This improvement reflects the capability of IRDFusion to progressively refine cross-modal features: the MFRM consolidates cross-modal information, and the DFFM emphasizes complementary differences, enabling the model to effectively suppress noise and enhance discriminative cues, which is particularly beneficial in complex multi-spectral scenarios.

\subsection{Ablation studies}
\label{subsec4.4}
\subsubsection{Different Modules}
\label{subsubsec4.4.1}
\begin{table}[!t]
\caption{Effects of different modules}
\resizebox{\columnwidth}{!}{%
\begin{tabular}{@{}cc|cccc|cccc|cccc@{}}
\toprule
\multirow{2}{*}{MFRM} & \multirow{2}{*}{DFFM} & \multicolumn{4}{c|}{mAP50}                                                                                                                                                                                                    & \multicolumn{4}{c|}{mAP75}                                                                                                                                                                                                    & \multicolumn{4}{c}{mAP}                                                                                                                                                                                                       \\ \cmidrule(l){3-14} 
                     &                       & person                                                & car                                                   & bicycle                                               & all                                                   & person                                                & car                                                   & bicycle                                               & all                                                   & person                                                & car                                                   & bicycle                                               & all                                                   \\ \midrule
                     & \multicolumn{1}{r|}{} & 89.1                                                  & 92.4                                                  & 72.9                                                  & 84.8                                                  & 46.2                                                  & 70.4                                                  & 15.5                                                  & 44.0                                                  & 48.5                                                  & 62.5                                                  & 28.7                                                  & 46.9                                                  \\  \midrule
\Checkmark                   & \multicolumn{1}{r|}{} & \begin{tabular}[c]{@{}c@{}}89.4\\ (+0.3)\end{tabular} & \begin{tabular}[c]{@{}c@{}}92.2\\ (-0.2)\end{tabular} & \begin{tabular}[c]{@{}c@{}}77.3\\ (+4.4)\end{tabular} & \begin{tabular}[c]{@{}c@{}}86.3\\ (+1.5)\end{tabular} & \begin{tabular}[c]{@{}c@{}}48.1\\ (+1.9)\end{tabular} & \begin{tabular}[c]{@{}c@{}}66.5\\ (-3.9)\end{tabular} & \begin{tabular}[c]{@{}c@{}}23.2\\ (+7.7)\end{tabular} & \begin{tabular}[c]{@{}c@{}}46.0\\ (+2.0)\end{tabular} & \begin{tabular}[c]{@{}c@{}}49.3\\ (+0.8)\end{tabular} & \begin{tabular}[c]{@{}c@{}}61.2\\ (-1.3)\end{tabular} & \begin{tabular}[c]{@{}c@{}}35.4\\ (+6.7)\end{tabular} & \begin{tabular}[c]{@{}c@{}}48.6\\ (+1.7)\end{tabular} \\ \midrule
                     & \Checkmark                      & \begin{tabular}[c]{@{}c@{}}89.7\\ (+0.6)\end{tabular} & \begin{tabular}[c]{@{}c@{}}93.2\\ (+1.0)\end{tabular} & \begin{tabular}[c]{@{}c@{}}79.5\\ (+6.6)\end{tabular} & \begin{tabular}[c]{@{}c@{}}87.5\\ (+2.7)\end{tabular} & \begin{tabular}[c]{@{}c@{}}48.1\\ (+1.9)\end{tabular} & \begin{tabular}[c]{@{}c@{}}70.7\\ (+0.3)\end{tabular} & \begin{tabular}[c]{@{}c@{}}19.2\\ (+3.7)\end{tabular} & \begin{tabular}[c]{@{}c@{}}46.0\\ (+2.0)\end{tabular} & \begin{tabular}[c]{@{}c@{}}49.5\\ (+1.0)\end{tabular} & \begin{tabular}[c]{@{}c@{}}64.2\\ (+1.7)\end{tabular} & \begin{tabular}[c]{@{}c@{}}33.9\\ (+5.2)\end{tabular} & \begin{tabular}[c]{@{}c@{}}49.2\\ (+2.3)\end{tabular} \\ \midrule
\Checkmark                    & \Checkmark                      & \begin{tabular}[c]{@{}c@{}}89.4\\ (+0.3)\end{tabular} & \begin{tabular}[c]{@{}c@{}}92.6\\ (+0.2)\end{tabular} & \begin{tabular}[c]{@{}c@{}}82.8\\ (+9.9)\end{tabular} & \begin{tabular}[c]{@{}c@{}}88.3\\ (+3.5)\end{tabular} & \begin{tabular}[c]{@{}c@{}}54.8\\ (+8.6)\end{tabular} & \begin{tabular}[c]{@{}c@{}}67.6\\ (-2.8)\end{tabular} & \begin{tabular}[c]{@{}c@{}}21.7\\ (+6.2)\end{tabular} & \begin{tabular}[c]{@{}c@{}}48.0\\ (+4.0)\end{tabular} & \begin{tabular}[c]{@{}c@{}}52.6\\ (+4.1)\end{tabular} & \begin{tabular}[c]{@{}c@{}}62.2\\ (-0.3)\end{tabular} & \begin{tabular}[c]{@{}c@{}}37.5\\ (+8.8)\end{tabular} & \begin{tabular}[c]{@{}c@{}}50.7\\ (+3.8)\end{tabular} \\ \bottomrule
\end{tabular}%
}
\label{module_ablation}
\end{table}

We conducted ablation studies using CoDetr with NiNfusion as the baseline (first row in \Cref{module_ablation}) to evaluate the contributions of the proposed modules.

\textbf{MFRM Module:} Introducing the MFRM module led to substantial improvements in all metrics for the "bicycle" and "person" categories, with the most pronounced gains observed for "bicycle". This suggests that features corresponding to "bicycle" are not prominent in either modality individually; by amplifying and fusing complementary features from both modalities, MFRM significantly enhances their representation, thereby improving detection performance. Notably, the growth rate for the "bicycle" category is the largest among all classes.

However, the "car" category exhibited a decline across all metrics. Detailed analysis indicates that while MFRM amplifies cross-modal features, it also inadvertently strengthens some irrelevant cross-modal features. In particular, the RGB modality contains background edges near object contours that are highly similar to contours in the infrared modality. During fusion, MFRM further amplifies these irrelevant edges, leading to reduced IoU and degraded detection performance for cars. This negative effect is especially pronounced at higher IoU thresholds, suggesting that in scenarios requiring precise localization, amplification of irrelevant features can become a limiting factor. Despite this localized drawback, overall metrics show clear improvements: mAP50, mAP75, and mAP increase by 1.5\%, 2.0\%, 1.7\%, respectively, demonstrating the overall efficacy of MFRM in enhancing detection across most categories.

\textbf{DFFM Module:} Upon incorporating the DFFM module, all metrics substantial gains overall, despite slight declines in specific categories, confirming that differential feature feedback effectively guides the fusion of relevant cross-modal features. The "bicycle" category again shows the largest gains, with mAP50, mAP75 and mAP increasing by 9.9\%, 6.2\%, 8.8\%, respectively, while the "person" category improves by 0.3\%, 8.6\%, 4.1\%. These results indicate that DFFM is particularly effective for objects that are visually blurred or weakly represented in both modalities, as it selectively reinforces critical features while suppressing redundant information. 

\textbf{ MFRM + DFFM module}: In summary, the ablation study clearly demonstrates that both MFRM and DFFM modules play complementary and crucial roles in enhancing cross-modal feature fusion. MFRM primarily amplifies and consolidates weak but informative features, while DFFM guides the fusion process to emphasize relevant features and suppress interference, together achieving significant improvements in detection performance across challenging categories. The combination of two modules leads to overall improvements of 3.5\%, 4.0\%, 3.8\% across mAP50, mAP75 and mAP respectively.
\subsubsection{Iteration Number}
\label{subsubsec4.4.2}
\begin{table}[!t]
\caption{Effects of different iteration Numbers}
\resizebox{\columnwidth}{!}{%
\begin{tabular}{@{}c|cccc|cccc|cccc@{}}
\toprule
\multirow{2}{*}{\begin{tabular}[c]{@{}c@{}}Iter\\ Num\end{tabular}} & \multicolumn{4}{c|}{mAP50}     & \multicolumn{4}{c|}{mAP75}    & \multicolumn{4}{c}{mAP}          \\ \cmidrule(l){2-13} 
                                                                             & person           & car           & bicycle       & all           & person        & car           & bicycle       & all           & person        & car           & bicycle       & all  \\ \midrule
Baseline                                                                     & 89.1             & 92.4          & 72.9          & 84.8          & 46.2          & 70.4          & 15.5          & 44            & 48.5          & 62.5          & 28.7          & 46.9 \\
1                                                                            & 89.0             & 92.8          & 74.5          & 85.4          & 45.5          & 69.5          & 12.9          & 42.6          & 48.1          & 63.2          & 28.4          & 46.6 \\
2                                                                            & 89.0             & \textbf{93.0} & 75.3          & 85.8          & 48.2          & \textbf{70.7} & 15.7          & 44.9          & 49.3          & 63.8          & 30.9          & 48.0 \\
3                                                                            & 88.8             & \textbf{93.0} & 77.2          & 86.3          & 46.3          & 70.6          & 18.0          & 45            & 48.2          & \textbf{63.9} & 32.1          & 48.1 \\
4                                                                            & \textbf{89.4}    & 92.6          & \textbf{82.8} & \textbf{88.3} & \textbf{54.8} & 67.6          & \textbf{21.7} & \textbf{48.0} & \textbf{52.6} & 62.2          & \textbf{37.5} & \textbf{50.7} \\
5                                                                            & 86.9             & 91.3          & 80.4          & 86.2          & 46.4          & 66.6          & 19.9          & 44.3          & 47.4          & 61.2          & 35.2          & 47.9 \\
6                                                                            & 88.8             & 91.7          & 76.5          & 85.7          & 48.8          & 67.9          & 17.2          & 44.7          & 49.6          & 62.3          & 31.1          & 47.7 \\ \bottomrule
\end{tabular}%
\label{iterative counts}
}
\end{table}
\begin{figure}[!t]
    \centering
    \includegraphics[width=1.0\linewidth]{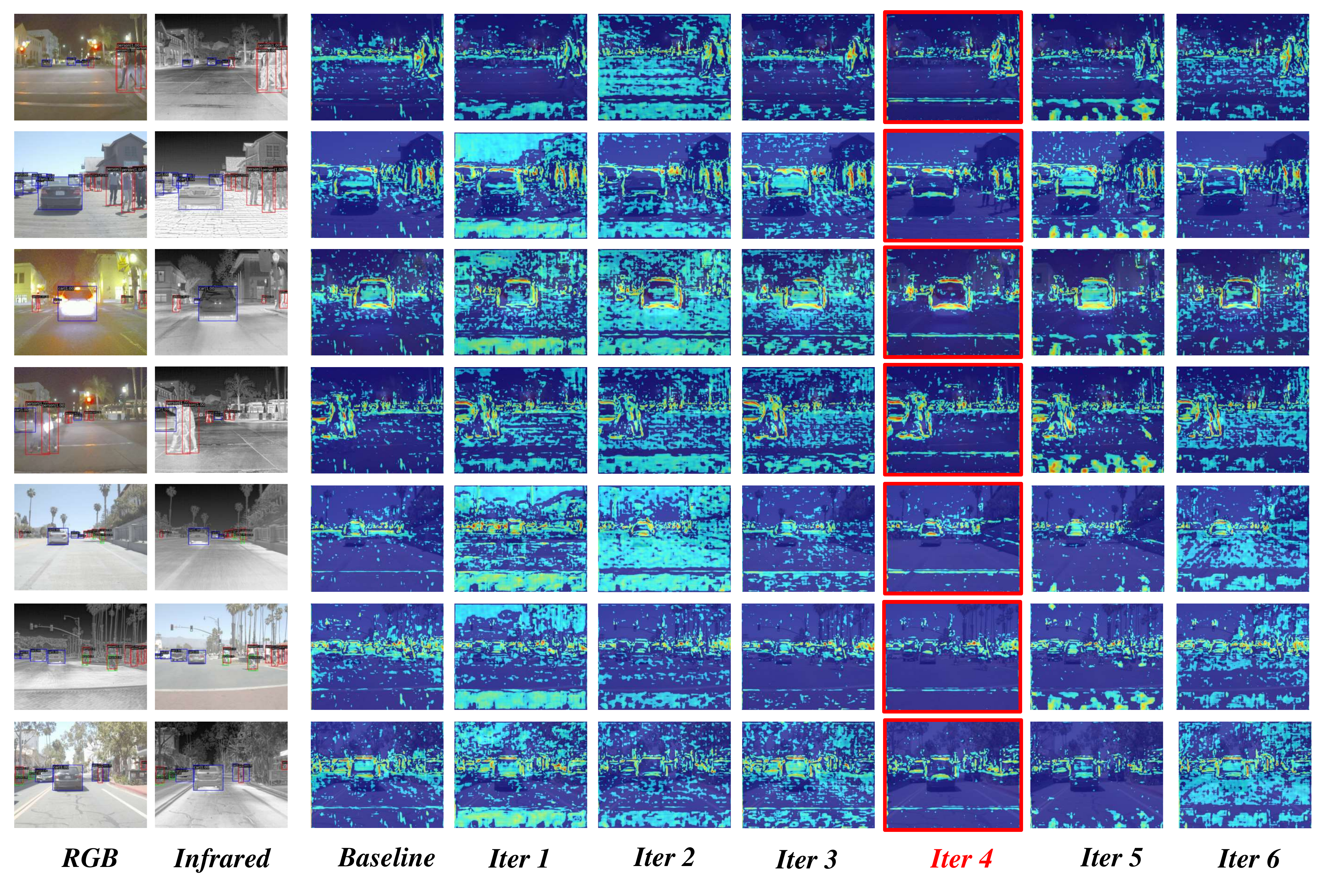}
    \caption{Visualization of different iterations of IRDFusion. The red markers indicate the optimal iteration number.}
    \label{heatmap}
\end{figure}

We have conducted ablation experiments to investigate the effect of the number of iterations in the DFFM module, with results summarized in \Cref{iterative counts}. Overall, the model achieves optimal performance at the fourth iteration in terms of mAP50. Although certain categories reach their peak performance at different iteration counts, these variations are minor, supporting the selection of four iterations as the standard setting for all experiments.

We also provide the visualization of fused feature maps of different iterations in \Cref{heatmap}. It is clear to observe that, when the number of iterations is too low, some redundant features are not fully suppressed. Limited feedback on differential features at lower iteration prevents effective elimination of redundant or noisy information within both modalities, reducing the model’s ability to integrate meaningful cross-modal features. By the fourth iteration, the DFFM module sufficiently removes irrelevant information while maximizing the fusion of critical features, resulting in optimal detection performance.

Conversely, we also find that further increasing the number of iterations leads to performance degradation. With excessive iterations, the differential features become progressively weakened, and interactions between modalities can introduce adverse effects. In particular, the amplification of background noise increases, which interferes with accurate feature integration. We attribute this decline to the model overemphasizing differential features at the expense of effectively consolidating features, thereby undermining overall fusion quality.

In summary, this ablation study demonstrates that an appropriate number of iterations is crucial for balancing the suppression of irrelevant features and the integration of meaningful cross-modal information. The fourth iteration achieves this balance, ensuring robust and effective performance across categories.

\subsubsection{Ablation study of single-branch detection}
\begin{table}[!t]
\caption{The individual detection results of different branches.}
\resizebox{\columnwidth}{!}{%
\begin{tabular}{@{}cc|cccc|cccc|cccc@{}}
\toprule
\multirow{2}{*}{Method}    & \multirow{2}{*}{Output} & \multicolumn{4}{c|}{mAP50}                                                                                                                                                                                                    & \multicolumn{4}{c|}{mAP75}                                                                                                                                                                                                    & \multicolumn{4}{c}{mAP}                                                                                                                                                                                                       \\ \cmidrule(l){3-14} 
                           &                         & person                                                & car                                                   & bicycle                                               & all                                                   & person                                                & car                                                   & bicycle                                               & all                                                   & person                                                & car                                                   & bicycle                                               & all                                                   \\ \midrule
\multirow{3}{*}{Baseline}  & RGB                     & 76.8                                                  & 86.3                                                  & 61.0                                                  & 74.7                                                  & 25.4                                                  & 51.9                                                  & 10.7                                                  & 29.3                                                  & 34.3                                                  & 50.9                                                  & 22.7                                                  & 36.0                                                  \\
                           & IR                      & 88.9                                                  & 93.0                                                  & 70.4                                                  & 84.1                                                  & 53.0                                                  & 68.6                                                  & 15.1                                                  & 45.6                                                  & 51.8                                                  & 62.9                                                  & 28.2                                                  & 47.6                                                  \\
                           & RGB+IR                  & 89.1                                                  & 92.4                                                  & 72.9                                                  & 84.8                                                  & 46.2                                                  & 70.4                                                  & 15.5                                                  & 44.0                                                  & 48.5                                                  & 62.5                                                  & 28.7                                                  & 46.9                                                  \\ \midrule
\multirow{6}{*}{IRDFusion} & RGB                     & \begin{tabular}[c]{@{}c@{}}78.5\\ (+1.7)\end{tabular} & \begin{tabular}[c]{@{}c@{}}87.1\\ (+0.8)\end{tabular} & \begin{tabular}[c]{@{}c@{}}66.2\\ (+5.2)\end{tabular} & \begin{tabular}[c]{@{}c@{}}77.2\\ (+2.5)\end{tabular} & \begin{tabular}[c]{@{}c@{}}29.4\\ (+4.0)\end{tabular} & \begin{tabular}[c]{@{}c@{}}55.5\\ (+3.6)\end{tabular} & \begin{tabular}[c]{@{}c@{}}14.4\\ (+3.7)\end{tabular} & \begin{tabular}[c]{@{}c@{}}33.1\\ (+3.8)\end{tabular} & \begin{tabular}[c]{@{}c@{}}37\\ (+2.7)\end{tabular}   & \begin{tabular}[c]{@{}c@{}}53.3\\ (+2.4)\end{tabular} & \begin{tabular}[c]{@{}c@{}}25.9\\ (+3.2)\end{tabular} & \begin{tabular}[c]{@{}c@{}}38.7\\ (+2.7)\end{tabular} \\  \cmidrule(l){2-14}
                           & IR                      & \begin{tabular}[c]{@{}c@{}}89.5\\ (+0.6)\end{tabular} & \begin{tabular}[c]{@{}c@{}}93.0\\ (+0.0)\end{tabular} & \begin{tabular}[c]{@{}c@{}}78.2\\ (+7.8)\end{tabular} & \begin{tabular}[c]{@{}c@{}}86.9\\ (+2.8)\end{tabular} & \begin{tabular}[c]{@{}c@{}}52.4\\ (-0.6)\end{tabular} & \begin{tabular}[c]{@{}c@{}}70.3\\ (+1.7)\end{tabular} & \begin{tabular}[c]{@{}c@{}}14.7\\ -(0.4)\end{tabular} & \begin{tabular}[c]{@{}c@{}}45.8\\ (+0.2)\end{tabular} & \begin{tabular}[c]{@{}c@{}}51.7\\ (-0.1)\end{tabular} & \begin{tabular}[c]{@{}c@{}}63.5\\ (+0.6)\end{tabular} & \begin{tabular}[c]{@{}c@{}}29.9\\ (+1.7)\end{tabular} & \begin{tabular}[c]{@{}c@{}}48.4\\ (+0.8)\end{tabular} \\ \cmidrule(l){2-14}
                           & RGB+IR                  & \begin{tabular}[c]{@{}c@{}}89.4\\ (+0.3)\end{tabular} & \begin{tabular}[c]{@{}c@{}}92.6\\ (+0.2)\end{tabular} & \begin{tabular}[c]{@{}c@{}}82.8\\ (+9.9)\end{tabular} & \begin{tabular}[c]{@{}c@{}}88.3\\ (+3.5)\end{tabular} & \begin{tabular}[c]{@{}c@{}}54.8\\ (+8.6)\end{tabular} & \begin{tabular}[c]{@{}c@{}}67.6\\ (-2.8)\end{tabular} & \begin{tabular}[c]{@{}c@{}}21.7\\ (+6.2)\end{tabular} & \begin{tabular}[c]{@{}c@{}}48.0\\ (+4.0)\end{tabular} & \begin{tabular}[c]{@{}c@{}}52.6\\ (+4.1)\end{tabular} & \begin{tabular}[c]{@{}c@{}}62.2\\ (-0.3)\end{tabular} & \begin{tabular}[c]{@{}c@{}}37.5\\ (+8.8)\end{tabular} & \begin{tabular}[c]{@{}c@{}}50.7\\ (+3.8)\end{tabular} \\ \bottomrule
\end{tabular}%
\label{branch ablation}
}
\end{table}
We also conducted ablation experiments to evaluate the contributions of each branch after integrating the IRDFusion module, as shown in \Cref{branch ablation}. In the RGB branch, incorporating infrared features through IRDFusion leads to improvements across all metrics compared to the baseline that relies solely on RGB features. Similarly, in the IR branch, fusing RGB features via IRDFusion yields notable gains in most metrics. These results demonstrate that both single-modality branches benefit significantly from cross-modal feature integration, with the independent IR branch generally outperforms the RGB branch.

For specific categories such as "bicycle," the improvements are particularly pronounced. This is because the features of "bicycle" are relatively indistinct in either RGB or IR modalities alone, and only through cross-modal fusion can salient object cues be sufficiently enhanced for reliable detection. Conversely, for certain "person" and "bicycle" instances, slight decreases in mAP75 and overall mAP are observed, likely due to interference from background clutter in the RGB modality, which may introduce minor noise into the fused features.

Finally, integrating IRDFusion across both branches achieves the best overall performance, confirming that joint fusion of RGB and IR modalities produces the most robust results. However, for the "car" category, a minor decline in mAP75 and overall mAP is observed, even compared to the IR branch alone. This is attributed to relatively clear car contours in the IR modality but blurred RGB features with nearby irrelevant background, causing the fusion to introduce slight misalignment and reduce IoU. Despite these localized declines, their overall impact on the model’s performance is negligible, underscoring the effectiveness of IRDFusion in enhancing cross-modal feature representation through iterative differential guidance.
\subsection{Different detection frameworks}
\label{subsubsec4.5}

\begin{table}[!t]
\caption{Comparison on different detection frameworks}
\resizebox{\columnwidth}{!}{%
\begin{tabular}{@{}ccc|ccc|ccc|ccc@{}}
\toprule
\multirow{2}{*}{Detector} & \multirow{2}{*}{Backbone}   & \multirow{2}{*}{Method}    & \multicolumn{3}{c|}{FLIR}  & \multicolumn{3}{c|}{LLVIP} & \multicolumn{3}{c}{M$^3$FD}   \\ \cmidrule(l){4-12} 
                          &                             &                            & mAP50 & mAP75  & mAP       & mAP50 & mAP75  & mAP       & mAP50 & mAP75  & mAP        \\ \midrule
\multirow{2}{*}{YOLOv5}   & \multirow{2}{*}{CSPDarknet} & Baseline                   & 79.9  & 34.6   & 40.2      & 96.8  & 71.2   & 62.7      & 89.1  & 65.8   & 59.8       \\ 
                          &                             & Ours                       & 84.8  & 38.1   & 43.4      & 97.9  & 75.7   & 65.5      & 89.8  & 65.8   & 60.5       \\ \midrule
\multirow{2}{*}{CoDetr}   & \multirow{2}{*}{ViT}        & Baseline                   & 84.8  & 44.0   & 46.9      & 98.0  & 80.7   & 69.5      & 87.1  & 61.1   & 58.2       \\
                          &                             & Ours                       & 88.3  & 48.0   & 50.7      & 98.4  & 83.1   & 70.9      & 90.8  & 65.4   & 61.9       \\ \bottomrule
\end{tabular}%
\label{dif_frameworks}
}
\end{table}

We further evaluated the adaptability of the proposed IRDFusion module by integrating it into different detection frameworks. First, in the anchor-based YOLOv5 framework, with NiNfusion as the baseline, experiments were conducted on the FLIR, LLVIP, and M$^3$FD datasets, as shown in \Cref{dif_frameworks}. Our method achieves consistent improvements across all datasets and metrics. Specifically, on FLIR, all three metrics increase by over 3\%; on LLVIP, the improvement in mAP75 is particularly notable, indicating enhanced precision in pedestrian localization; and on M$^3$FD, while the gains are smaller, there is still measurable improvement. Although YOLOv5 is designed for real-time applications and emphasizes speed, the integration of IRDFusion demonstrates that its feature fusion mechanism can still provide significant accuracy gains.

Similarly, in the CoDETR framework, a DETR variant, IRDFusion achieves substantial improvements, particularly on FLIR. On LLVIP, the most pronounced gain is observed in mAP75, suggesting that IRDFusion effectively enhances high-precision localization of pedestrians. On M$^3$FD, mAP50 shows a notable increase, with slight improvements in mAP75 and overall mAP. These results indicate that the iterative feedback mechanism of IRDFusion, through MFRM and DFFM, consistently strengthens cross-modal feature alignment and discriminability, regardless of the underlying detection architecture.

Collectively, these experiments demonstrate that IRDFusion is highly generalizable and robust across different frameworks and datasets. As a plug-and-play module, it can be seamlessly integrated into both speed-oriented frameworks like YOLOv5 and performance-oriented frameworks like CoDETR, consistently enhancing detection performance through adaptive cross-modal fusion.
\subsection{Comparison of Alternative Methods of Replacement}
\label{subsubsec4.6}
\begin{table}[!t]
\caption{Comparison with different attention methods}
\resizebox{\columnwidth}{!}{%
\begin{tabular}{@{}c|cccc|cccc|cccc@{}}
\toprule
          & \multicolumn{4}{c|}{mAP50}     & \multicolumn{4}{c|}{mAP75}     & \multicolumn{4}{c}{mAP}        \\ \cmidrule(l){2-13} 
          & person & car  & bicycle & all  & person & car  & bicycle & all  & person & car  & bicycle & all  \\ \midrule
Self-Attention&88.6&91.9  &79.4     &86.6  &49.9    &67.2  &31.4     &49.5  &50.2    &62.0  &38.5     &50.2  \\
Cross-Attention & 90.4   & 93.4 & 76.5    & 86.8 & 51.6   & 71.9 & 22.4    & 48.6 & 51.3   & 65.0 & 34.7    & 50.3 \\
IRDFusion & 89.4   & 92.6 & 82.8    & 88.3 & 54.8   & 67.6 & 21.7    & 48.0 & 52.6   & 62.2 & 37.5    & 50.7 \\ \bottomrule
\end{tabular}
\label{alternative methods comparsion}}
\end{table}
To further validate the effectiveness of the proposed fusion design, we conducted comparative experiments by replacing the MFRM modules with alternative mechanisms, including Cross-Attention in ICAFusion \cite{SHEN2024109913} and standard self-attention \cite{vaswani2017attention} in Transformer. The results indicate that, although these alternatives offer certain advantages, they still underperform compared to IRDFusion. Specifically, When fed back into the MFRM, differential information requires both intra-modal consistency and cross-modal alignment. Using only self-attention or cross-attention fails to simultaneously satisfy these two requirements, leading to suboptimal performance. In contrast, IRDFusion combines MFRM and DFFM to simultaneously reinforce cross-modal features and dynamically extract differential cues. The iterative feedback mechanism amplifies complementary object information while suppressing redundant noise, leading to more stable, discriminative, and well-aligned feature representations. These results demonstrate that the unique design of IRDFusion is crucial for achieving superior cross-modal fusion performance compared to conventional attention-based alternatives.

\subsection{Visualization}
\label{subsec4.7}
\begin{figure}[!t]
    \centering
    \includegraphics[width=1.0\linewidth]{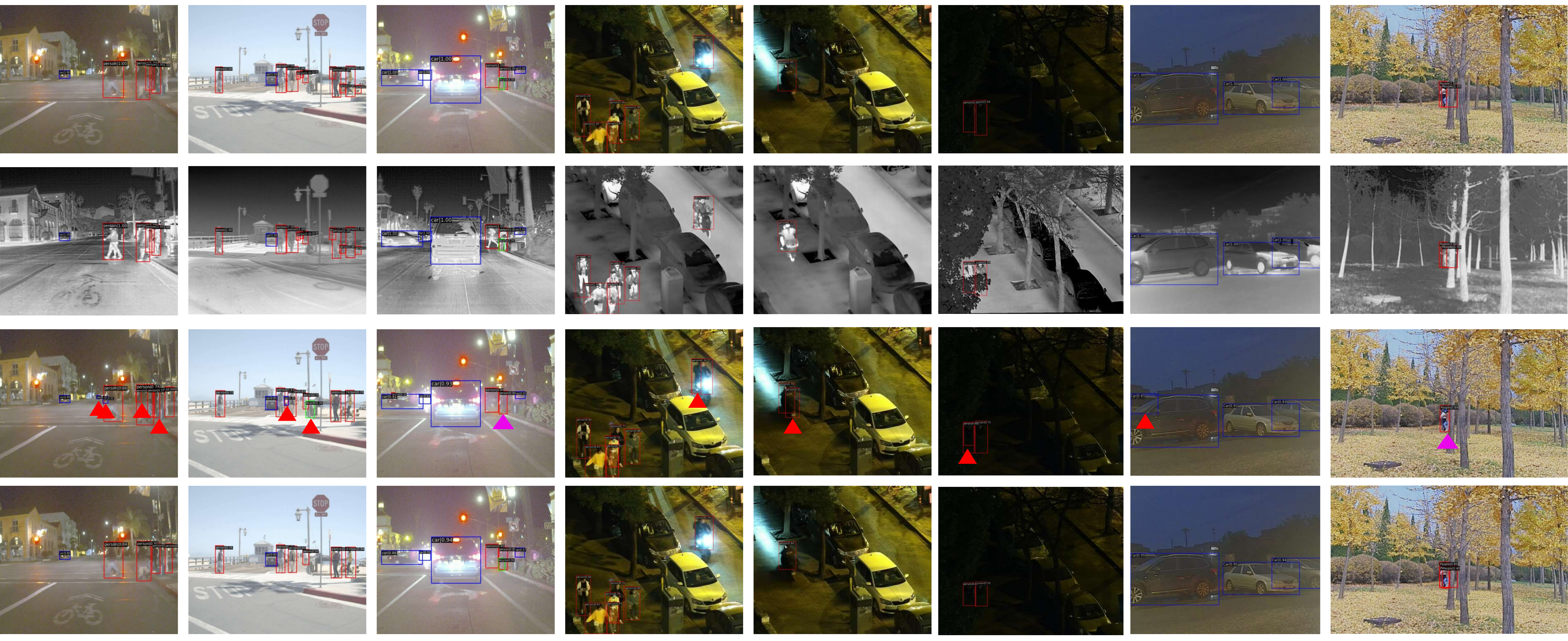}
    \caption{Sample detections of our method on FLIR, LLVIP, and M$^3$FD 
 datasets. The first and second rows are the input RGB and IR images, the third row is the detection result of the baseline, and the fourth row is the detection result of our method. The red triangle markers indicate \textcolor{red}{false detections} of the baseline, while the pink triangle markers indicate \textcolor[RGB]{235,5,235}{missed detections} of the baseline.}
    \label{demo}
\end{figure}

As shown in \Cref{demo}, the baseline method employs a global feature fusion strategy, which exhibits notable limitations in detection tasks. Specifically, false detections (red triangles) appear in columns 1, 2, 4, 5, 6, and 7, while missed detections (pink triangles) are observed in columns 3 and 8. These errors arise because the baseline fails to differentiate between cross-modal and differential features across modalities, relying instead on coarse global fusion. Such a strategy inadequately integrates complementary cues, leading to the omission or degradation of critical object information and, consequently, reduced overall detection performance.

In contrast, our proposed IRDFusion method explicitly separates cross-modal and differential features, and leverages both types in the fusion process. This fine-grained approach effectively suppresses redundant background information while amplifying complementary cues, significantly reducing both false and missed detections. The visualization results thus illustrate the capability of IRDFusion to produce more precise and discriminative cross-modal representations, further validating the importance of iterative feature fusion in enhancing multispectral object detection performance.
\subsection{Failure Case}
\label{subsec4.8}
\begin{figure}
    \centering
    \includegraphics[width=1.0\linewidth]{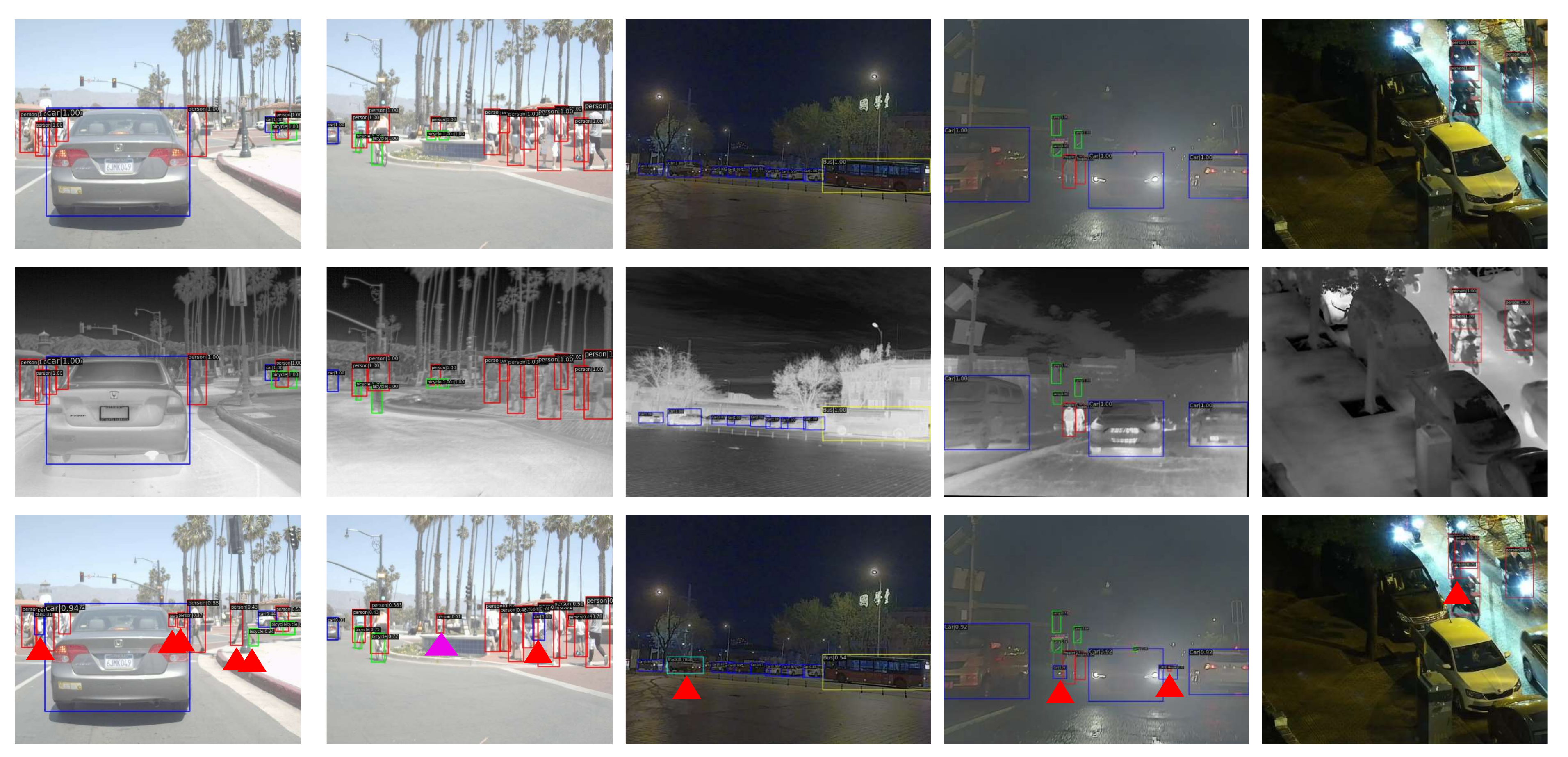}
    \caption{False detections of our method on FLIR, LLVIP, and M$^3$FD 
 datasets. The first and second rows are the input RGB and IR images, and the third row is the detection result of our method. The red triangle markers indicate \textcolor{red}{false detections} of the IRDFusion, while the pink triangle markers indicate \textcolor[RGB]{235,5,235}{missed detections} of the IRDFusion.}
    \label{falsecase}
\end{figure}
As shown in \Cref{falsecase}, IRDFusion exhibits limitations when detecting small, heavily occluded objects. Case studies indicate that under such conditions, both modal-shared features (e.g., contours, spatial layouts) and differential features (e.g., color or thermal contrasts) become less distinguishable. This ambiguity can reduce the effectiveness of the MFRM in accurately capturing cross-modal cues and hinder the DFFM from extracting complementary information via differential feedback. As a result, the fused feature representations are weakened, leading to occasional false positives and missed detections. Addressing these challenges may require future exploration of multi-scale feature enhancement strategies or dynamic deformable attention mechanisms, which could strengthen cross-modal feature interactions and improve detection performance under severe occlusion.
\subsection{Limitations}
\label{subsubsec4.8}
\begin{table}[!h]
\centering
\caption{Speed Comparison}
\resizebox{0.5\columnwidth}{!}{%
\begin{tabular}{@{}ccccc@{}}
\toprule
\multicolumn{1}{l}{Detector} & \multicolumn{1}{l}{Method} & \multicolumn{1}{l}{FPS} & \multicolumn{1}{l}{params(M)} & \multicolumn{1}{l}{GFLOP} \\ \midrule
\multirow{2}{*}{YOLOv5}      & Baseline                   & 45.3                    & 75.4                       & 91.0                     \\
                             & Ours                       & 17.0                    & 134.0                      & 122.8                    \\ \midrule
\multirow{2}{*}{CoDetr}      & Baseline                   & 3.6                     & 485.3                      & 944.9                    \\
                             & Ours                       & 3.1                     & 510.5                      & 1213.5                   \\ \bottomrule
\end{tabular}%
\label{speed}
}
\end{table}
According to the results in \Cref{speed}, it is evident that the detection speed of the YOLO framework significantly decreases after introducing the IRDFusion module. In the case of the CoDetr framework, although the reduction in speed is relatively smaller, there is a substantial increase in both the model's parameter count and GFLOPs. Specifically, the parameter count increases by approximately 60M in the YOLO framework and about 25M in the CoDetr framework. Meanwhile, the FPS drops significantly. This could potentially be due to the iterative processes within the IRDFusion module. Given that CoDetr utilizes ViT as its backbone, and the baseline model already possesses a large number of parameters and high FLOPs, it demands considerable computational resources. Although IRDFusion improves detection accuracy, its high computational cost limits real-time deployment, especially on edge devices.

\section{Conclusion}
\label{sec5}
In this paper, we propose IRDFusion, a novel multispectral object detection framework that effectively integrates RGB and infrared modalities through progressive, fine-grained feature fusion. The framework is built upon two complementary modules: the Mutual Feature Refinement Module (MFRM), which enhances cross-modal semantic alignment and suppresses redundant background, and the Differential Feature Feedback Module (DFFM), which dynamically extracts inter-modal differential cues and iteratively feeds them back to guide fusion. By enhancing cross-modal cues and guiding differential features, IRDFusion leverages the MFRM to strengthen cross-modal semantic consistency and the DFFM to iteratively refine differential information, progressively amplifying salient object signals while suppressing common-mode noise, thereby producing highly discriminative and well-aligned feature representations.

Extensive experiments on FLIR, LLVIP, and M$^3$FD datasets, including ablation studies, cross-framework evaluations, and visualizations, demonstrate the robustness and effectiveness of IRDFusion. The method consistently outperforms state-of-the-art approaches, particularly in challenging conditions such as low illumination and complex backgrounds. Comparative studies replacing the MFRM/DFFM modules further validate the importance of our iterative differential fusion strategy. Visualization results indicate that IRDFusion, by leveraging cross-modal feature enhancement and iterative guidance from differential cues, effectively reduces false positives and missed detections. However, the method still exhibits limitations in detecting small or heavily occluded objects.

While IRDFusion achieves significant performance improvements, computational efficiency and real-time capability remain challenging for practical deployment. Future work will focus on lightweight optimization, multi-scale feature enhancement, and dynamic attention mechanisms to further improve efficiency and detection performance under severe occlusion. Overall, this work provides a robust and generalizable solution for multispectral object detection, highlighting the value of iterative differential-guided cross-modal fusion.
\section*{Acknowledgments}
This work was supported in part by the National Natural Science Foundation of China under Grant No. 61903164, 62173186 and in part by Natural Science Foundation of Jiangsu Province in China under Grants BK20191427. Heng Fan receives no financial support for the research, authorship, and/or publication of this article.

\bibliography{ref}
\end{document}